\documentclass{ws-rv961x669}
\usepackage[]{graphicx}
\usepackage{amsmath,amssymb,url}
\usepackage{ws-rv-van}
\usepackage{color}

\DeclareMathOperator*{\argmin}{arg\,min}

\begin{document}
\chapter[Cortical Coordinates]{3D Normal Coordinate Systems for Cortical Areas}

\author{J. Tilak Ratnanather}
\address{Center for Imaging Science \&  Institute for Computational Medicine, Department of Biomedical Engineering\\
Johns Hopkins University, Baltimore MD 21218\\
tilak@cis.jhu.edu}

\author{Sylvain Arguill\`ere}
\address{CNRS, Universit\'e Claude Bernard Lyon 1 and CNRS UMR 5208,\\ Institut Camille Jordan, Lyon, France\\
sarguillere@gmail.com}

\author{Kwame S. Kutten}
\address{Center for Imaging Science, Johns Hopkins University, Baltimore MD 21218\\
kkutten1@jhmi.edu}

\author{Peter Hubka}
\address{Institute of AudioNeuroTechnology \& Department of Experimental Otology, Hannover Medical School, Hannover, Germany\\
hubka.peter@mh-hannover.de}

\author{Andrej Kral}
\address{Institute of AudioNeuroTechnology \& Department of Experimental Otology, Hannover Medical School, Hannover, Germany\\
kral.andrej@mh-hannover.de}

\author{Laurent Younes}
\address{Center for Imaging Science \& Institute for Computational Medicine, Department of Applied Mathematics and Statistics\\
Johns Hopkins University, Baltimore MD 21218\\
younes@cis.jhu.edu}

\begin{abstract}
A surface-based diffeomorphic algorithm to generate 3D coordinate grids in the cortical ribbon is described. In the grid, normal coordinate lines are generated by the diffeomorphic evolution from the gray/white (inner) surface to the gray/csf (outer) surface. Specifically, the cortical ribbon is described by two triangulated surfaces with open boundaries. Conceptually, the inner surface sits on top of the white matter structure and the outer on top of the gray matter. It is assumed that the cortical ribbon consists of cortical columns which are orthogonal to the white matter surface. This might be viewed as a consequence of the development of the columns in the embryo. It is also assumed that the columns are orthogonal to the outer surface so that the resultant vector field is orthogonal to the evolving surface. Then the distance of the normal lines from the vector field such that the inner surface evolves diffeomorphically towards the outer one can be construed as a measure of thickness. Applications are described for the auditory cortices in human adults and cats with normal hearing or hearing loss. The approach offers great potential for cortical morphometry.
\end{abstract}

\body

\section{Introduction}
A conspicuous feature of the mammalian brain is the folded cortex, i.e. cortical ribbon, which maximises surface area within a confined space with the folding varying in degree from large mammals to small ones \citep{Welker1990}. The cortex consists of neural tissue called gray matter, containing mostly neuronal cell bodies and unmyelinated fibers. The cortex can be divided into several regions or areas. For example, the human cortex has about 180 cortical regions \citep{Glasser2016} that serve different functional roles \citep{Rakic1988}, vary in size and are connected via white matter containing axonal, usually myelinated, fibers. At the microscopic level, each cortical region is composed of fundamental units called cortical columns \citep{Mountcastle1957} that traverse vertically from the white matter to the surface just below the pia matter. The cortical column consists of six layers that are stacked horizontally on top of each other \citep{Cajal1899}. The orthogonality of cortical columns stems from the fact that in the embryonic stage, columns are formed by neurons migrating through the initially flat cortical plate which in turns folds as the brain expands \citep{Geschwind2013}.

Here cortical morphometry is more than volumetric analysis. The surface area of the interface with the white matter and thickness combine to yield the volume of the cortical cortex. Yet surface area and thickness can be differentially affected \citep{Winkler2018} in development and disease due to the influence of genes \citep{Rakic1988} and environment \citep{Rakic2009}. Expansion in surface area may be associated with increase in number of columns while increasing thickness may be associated with changes in the columnar microcircuitry \citep{Wagstyl2018}.

\begin{figure}
\begin{tabular}{cc}
\includegraphics[width=0.5\textwidth]{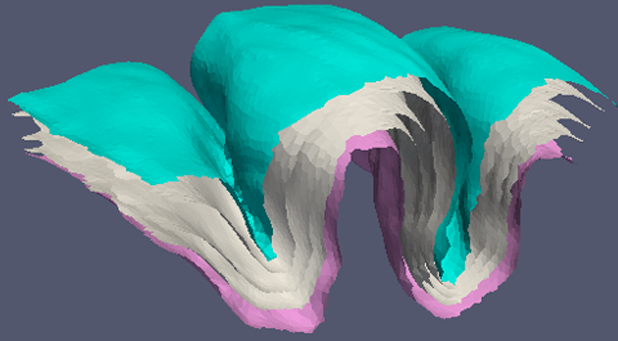} &
\includegraphics[width=0.25\textheight]{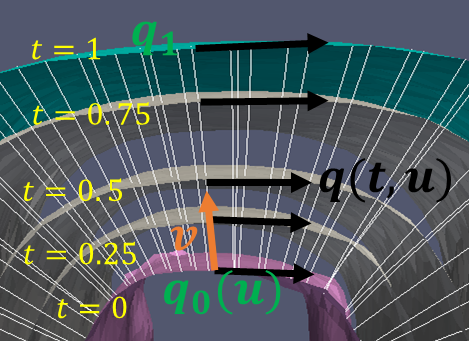}
\end{tabular}
\caption{Left: a cortical ribbon from a cat brain with a continuous deformation of the inner (pink) surface onto the outer (blue) surface described by the gray surfaces. Right: A 3D coordinate system for the cortical ribbon between the parametrized inner and outer surfaces ($q_0$ and $q_1$) is represented by the mapping $\Phi: U\times [0,1] \to \mathbb R^3$ where the coordinates $(u,t)$ of the inner surface are mapped to those of the evolving surface $q(t,u)$ indicated by three intermediate surfaces at $t=0.25, 0.5$ and $0.75$ and the vector field $v$ that is orthogonal to the evolving surface.  \label{Cartoon}}
\end{figure}

Morphometric measures such as surface area, thickness, volume and curvature warrant a precise 3D coordinate system that reflects the columnar and laminar structure of the cortical area. Therefore it is necessary to construct a coordinate system that traverses normally between the inner and outer boundaries of the cortical ribbon. In other words, a continuous deformation of the inner surface onto the outer surface, using large deformations to accommodate the highly folded cortical ribbon as shown in the left panel of Fig. \ref{Cartoon}. Such a mapping can be developed in the Large Deformation Diffeomorphic Metric Mapping (LDDMM) framework \citep{Beg2005} but with imposed normal constraints to enable the inner surface evolve to the outer surface. This follows but contrasts with orthogonal or curvilinear coordinate image-based systems proposed in recent years \citep{Leprince2015,Waehnert2014,Jones2000,Das2009}.

\section{Methods}
\subsection{Theory}
 The cortical region is endowed with normal coordinates defined by the apex and base of the cortical columns which in turn form natural coordinates for, i.e. parametrize, the inner and outer surface. Thus it is natural to view the columns tracing a diffeomorphism between two coordinate systems as illustrated in the right panel of Fig. \ref{Cartoon}. Consider two parameterized non-intersecting open surfaces: $q_0:U \rightarrow \mathbb{R}^3$ (the inner surface) and $q_1:U \rightarrow \mathbb{R}^3$ (the outer surface) where $U$ is an open subset of $\mathbb R^2$ and $q$ denotes a smooth embedding. The goal is to find a time-dependent surface parametrization $q(t):U \rightarrow \mathbb{R}^3$ such that (i) $q(0)=q_0$ and $q(1)$ is a reparametrization of the surface associated with $q_1$; (ii) $q(t)$ is obtained from $q(0)$ via a time-dependent diffeomorphism, so that $q(t,x) = \varphi(t,q_0(x))$ for some $\varphi(t,\cdot)$; (iii) the associated Eulerian velocity $\partial_tq(t,x)$  is at all times perpendicular to the evolving surface.

Let $V$ be a reproducing kernel Hilbert space (RKHS) of vector fields $v: \mathbb R^3 \to \mathbb R^3$, that is assumed to be, for some $p\geq 2$, continuously included in $C^p_0(\mathbb{R}^3, \mathbb R^3)$, the space of $C^p$ vector fields with all partial derivatives up to order $p$ tending to 0 at infinity, equipped with the usual supremum norm over all $x\in \mathbb R^d$ and all derivatives of order $p$ or less. Let $v \mapsto \|v\|_V$ denote the Hilbert norm on $V$. Consider the shape space $\mathcal M$ of all $C^1$ embeddings $q: U \to \mathbb R^3$, and assume that $q_0, q_1\in \mathcal M$. For $q\in \mathcal M$, let $S_q = q(U)$, which is, by assumption a submanifold of $\mathbb R^3$.

For $q\in \mathcal M$, let $V_q = \{v \in V: Dq(x)^tv(q(x)) = 0, x\in U\}$ denote the space of all vector fields in $V$ that are perpendicular to the evolving surface $S_q$. Also, define, for $v\in V$, the surface-dependent norm
\[
\|v\|_q^2 = \|v\|_V^2 + \lambda \int_{S_q} |Dv|^2 d\sigma_{S_q},
\]
where $Dv$ is the differential of $v$, $|Dv|^2$ is the squared Frobenius norm of $Dv$ (the sum of its squared entries) and $\sigma_{S_q}$ is the volume measure of $S_q$. Here $\|v\|_q$ is a hybrid norm\cite{younes2018hybrid} where the $V$ norm, which is the standard norm in LDDMM, ensures that the transformation is diffeomorphic and penalizes non-local collisions within surfaces and the second term restricts local deformations on the surfaces. Combining the two terms permits larger penalties for local perturbations and milder ones for non-local collisions, which is well adapted for highly folded surfaces such as the ones considered here. Finally, assume that the chosen ``data attachment'' term is represented as a function $q \mapsto D(S_q, S_{q_1})$, differentiable in $q$. Examples of such terms are provided by including the space of surfaces in the dual of some suitable RKHS, leading to representations as measures \citep{glaunes2004diffeomorphic}, currents \citep{Vaillant2005} or varifolds \citep{Charon2013}. Specifically, the latter is used with
\[
D(S,S') = \int_{S}\int_{S'} \chi(p, p') (1+(N_S(p)^tN_{S'}(p'))^2)\, d\sigma_S(p)\, d\sigma_{S'}(p')
\]
where $N_S$ and $N_{S'}$ are the unit normals to $S$ and $S'$, respectively and $\chi$ is a positive kernel. It is important that these data attachment terms depend only on the manifolds, $S$, $S'$ and not on their parameterization, to ensure a reparameterization-invariant registration method. This is true for $D$ because it is defined in terms of intrinsic properties of the surfaces embedded in $\mathbb R^3$.

With this notation, the following optimal control problem is solved. Minimize
\[
F(q(\cdot), v(\cdot)) = \int_0^1 \|v(t)\|_{q(t)}^2 \, dt + D(S_{q(1)}, q_1),
\]
subject to $q(0) = q_0$, $v(t) \in V_{q(t)}$ and $\partial_t q(t) = v(t, q(t))$. If $v(\cdot)$ is a solution, its flow $\varphi$, defined by $\partial_t \varphi(t) = v(t, \varphi(t))$, is a diffeomorphism such that $q(1) = \varphi(1, q_0)$. Computationally, the constraint $v\in V_q$ is enforced with an augmented Lagrangian \cite{nocedal2006nonlinear} which has been used in constrained LDDMM problems\cite{arguillere2014shape}.

The solution yields a 3D coordinate system for the space between $q_0$ and $q_1$ (cf. Fig. \ref{Cartoon}):
\[
\begin{aligned}[b]
\Phi: && U\times [0,1] &\to \mathbb R^3\\
&& (u, t) &\mapsto  q(t,u)
\end{aligned}\,.
\]
(Recall that $u\in U$ is a 2D parameter.) The surfaces $t = \mathrm{constant}$ can be interpreted as ``sheets'' between the inner and outer surfaces, and the transversal or normal lines $u = \mathrm{constant}$ as ``columnar lines''. By no means, does this geometric construction actually represent the laminar and columnar properties of the cortical region. However the lengths of the normal lines can be used as measures of the thickness of the cortical region with respect to $q_0(u)$.

\subsection{Data}
Three datasets were used as to illustrate the theory. In the following, the inner and outer surfaces were obtained from triangulations of segmented cortices \citep{Cheng2012}.
\subsubsection{Motor Cortex} Two MRI scans of a single subject with resolution $0.5\times0.5\times0.5$ mm/voxel taken a week apart were obtained from a 7T scanner (Philips Healthcare, The Netherlands). A subvolume of the primary motor cortex (M1) was manually segmented \citep{Mai2015} from which the anterior portion was used for the analysis.
\subsubsection{Primary and higher order auditory cortices in cats} MRI scans of two cats (one with hearing loss) with resolution $0.176\times0.176\times0.411$ mm/voxel were obtained from a 7T scanner (Bruker BioSpin, Germany). A subvolume encompassing the primary and higher order auditory cortices was segmented manually \citep{Reinoso-Suarez1961a}.
\subsubsection{Heschl's gyrus and Planum temporale in adults}  MRI scans of 10 adults (5 with profound hearing loss who consistently used Listening and Spoken Language with hearing aids after early diagnosis and intervention in infancy) were obtained from a 1.5T scanner (Phillips Healthcare, The Netherlands). Subvolumes encompassing Heschl's gyrus and Planum temporale were parcellated and segmented \citep{Ratnanather2003} with resolution $1\times1\times1$ mm/voxel.
\section{Results}
\begin{figure}
\begin{tabular}{cc}
\includegraphics[width=0.47\textwidth]{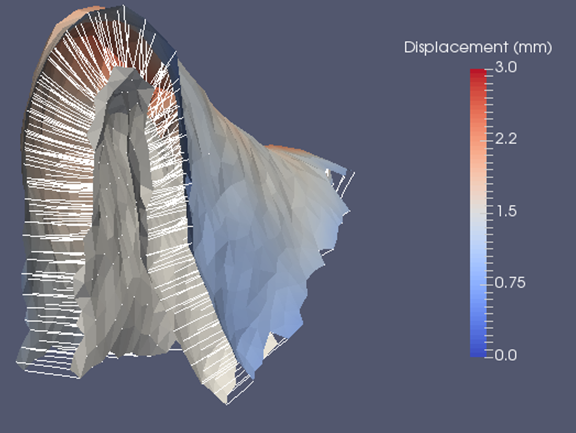} &
\includegraphics[width=0.47\textwidth]{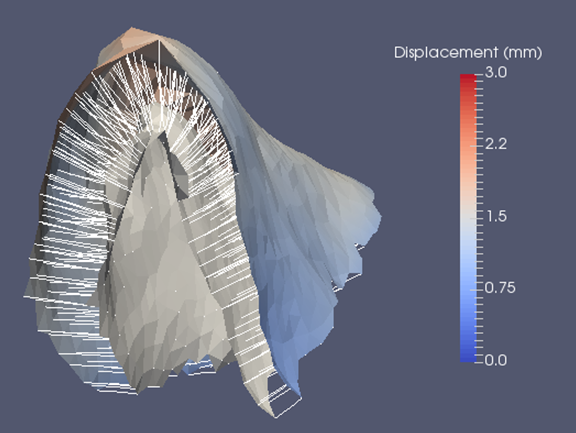}
\end{tabular}
\caption{Normal lines for the motor cortex in a single subject scanned one week apart. The color bar shows the variation in the lengths of the lines i.e a distribution of thickness.\label{M1}}
\end{figure}

Figure \ref{M1} shows the normal lines for the M1 cortex in a single subject scanned at two time points a week apart. The distribution of the lengths of the normal lines from the two scans (not shown) are virtually similar. So the method is robust to perturbations such as scan times, segmentations, triangulation and delineation.

\begin{figure}
\begin{tabular}{cc}
\includegraphics[width=0.47\textwidth]{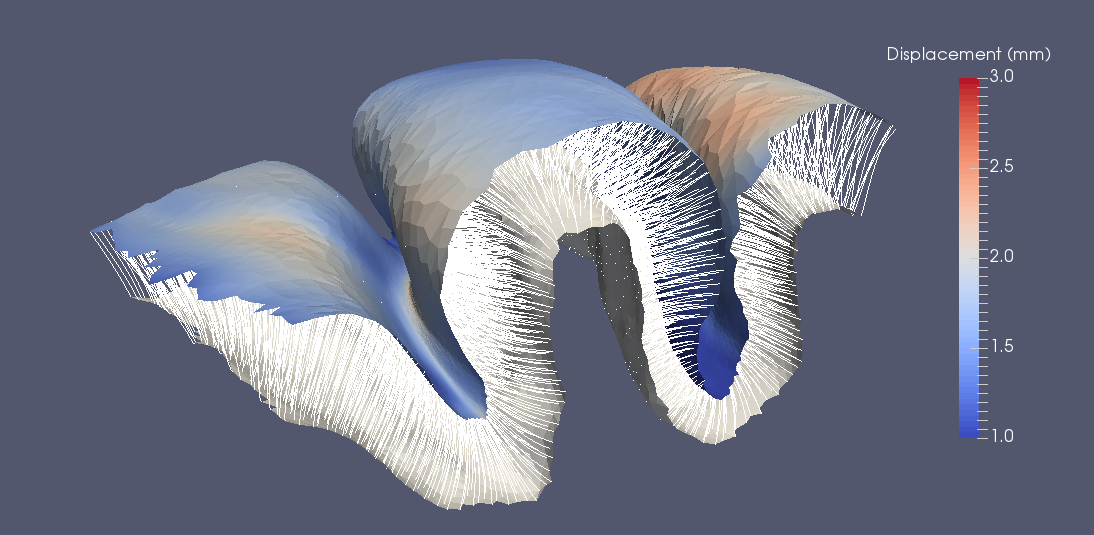} &
\includegraphics[width=0.47\textwidth]{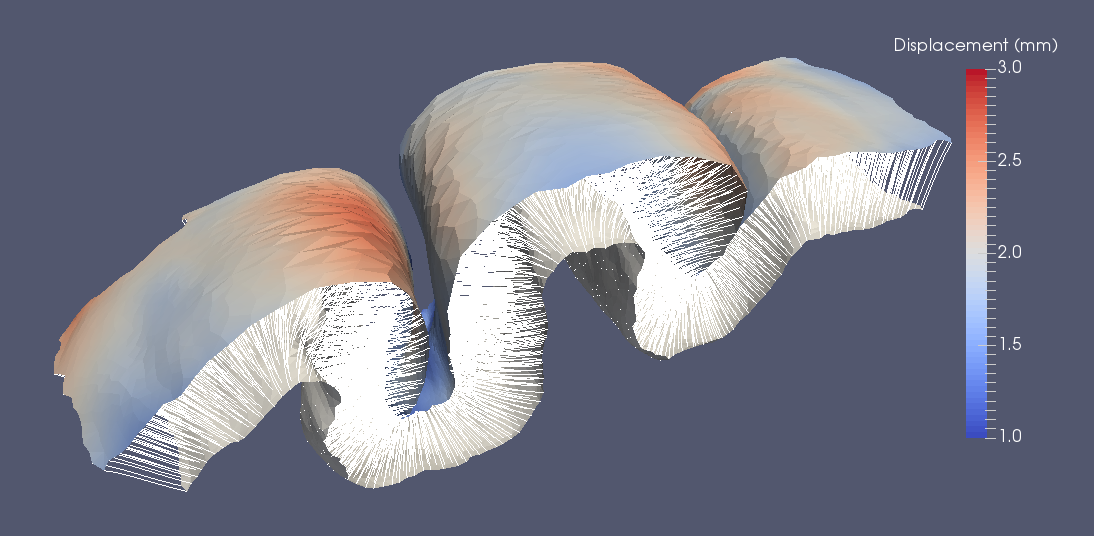} \\
\includegraphics[width=0.47\textwidth]{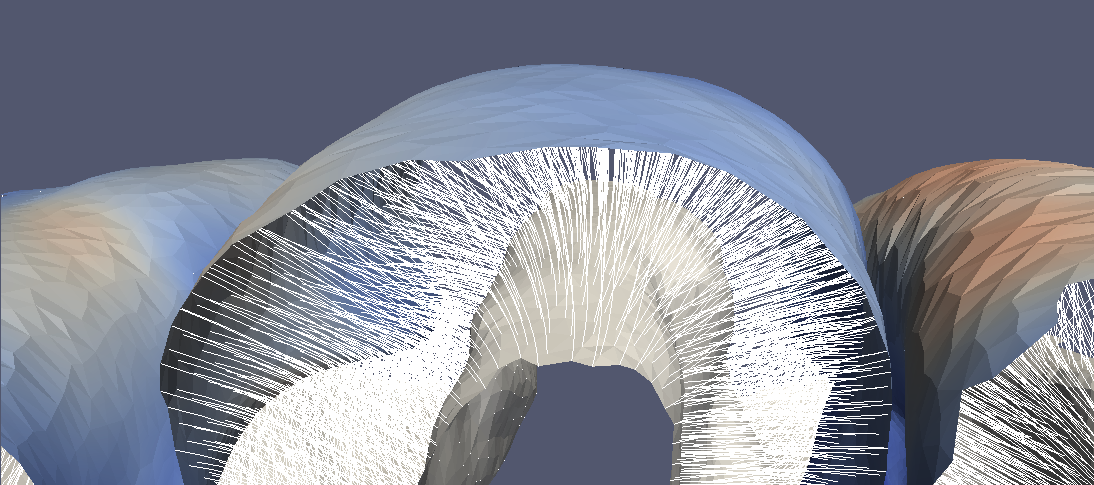} &
\includegraphics[width=0.47\textwidth]{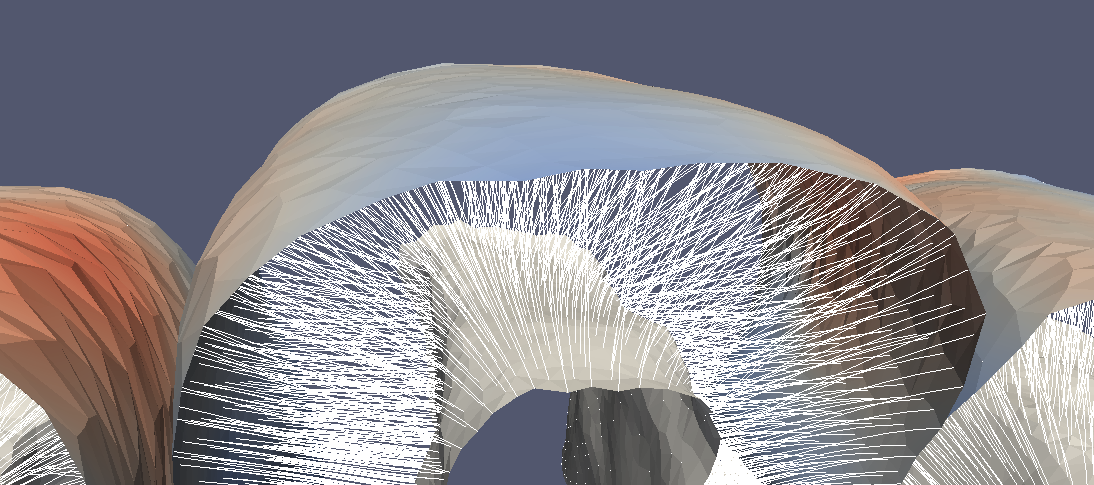} \\
\includegraphics[width=0.47\textwidth]{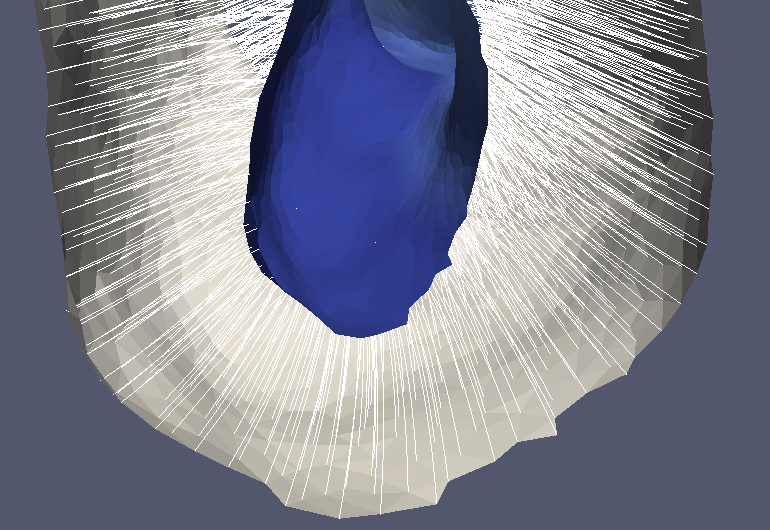} &
\includegraphics[width=0.47\textwidth]{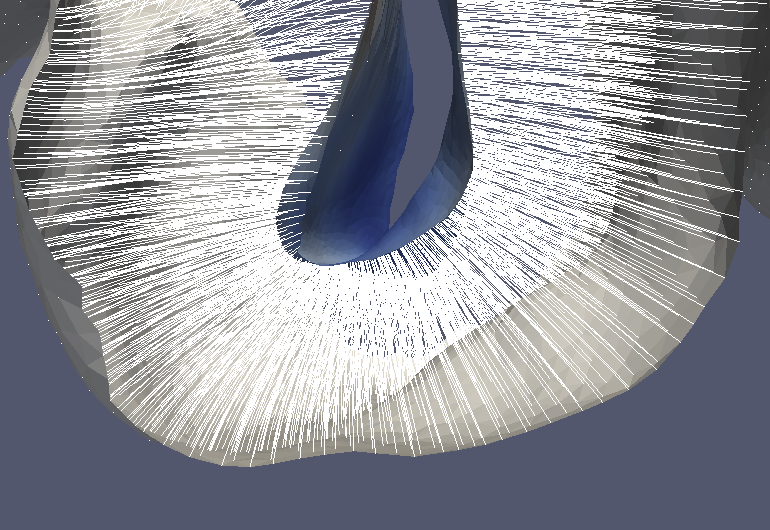}
\end{tabular}
\caption{Normal lines for primary and secondary auditory cortices in a normal hearing cat (left) and a congenitally deaf cat (right).\label{Cats}}
\end{figure}

Figure \ref{Cats} shows the results for one normal hearing cat and one congenitally deaf cat including zoomed views of one gyral crown and one sulcal fundus.

\begin{figure}
\begin{tabular}{cc}
\includegraphics[width=0.45\textwidth]{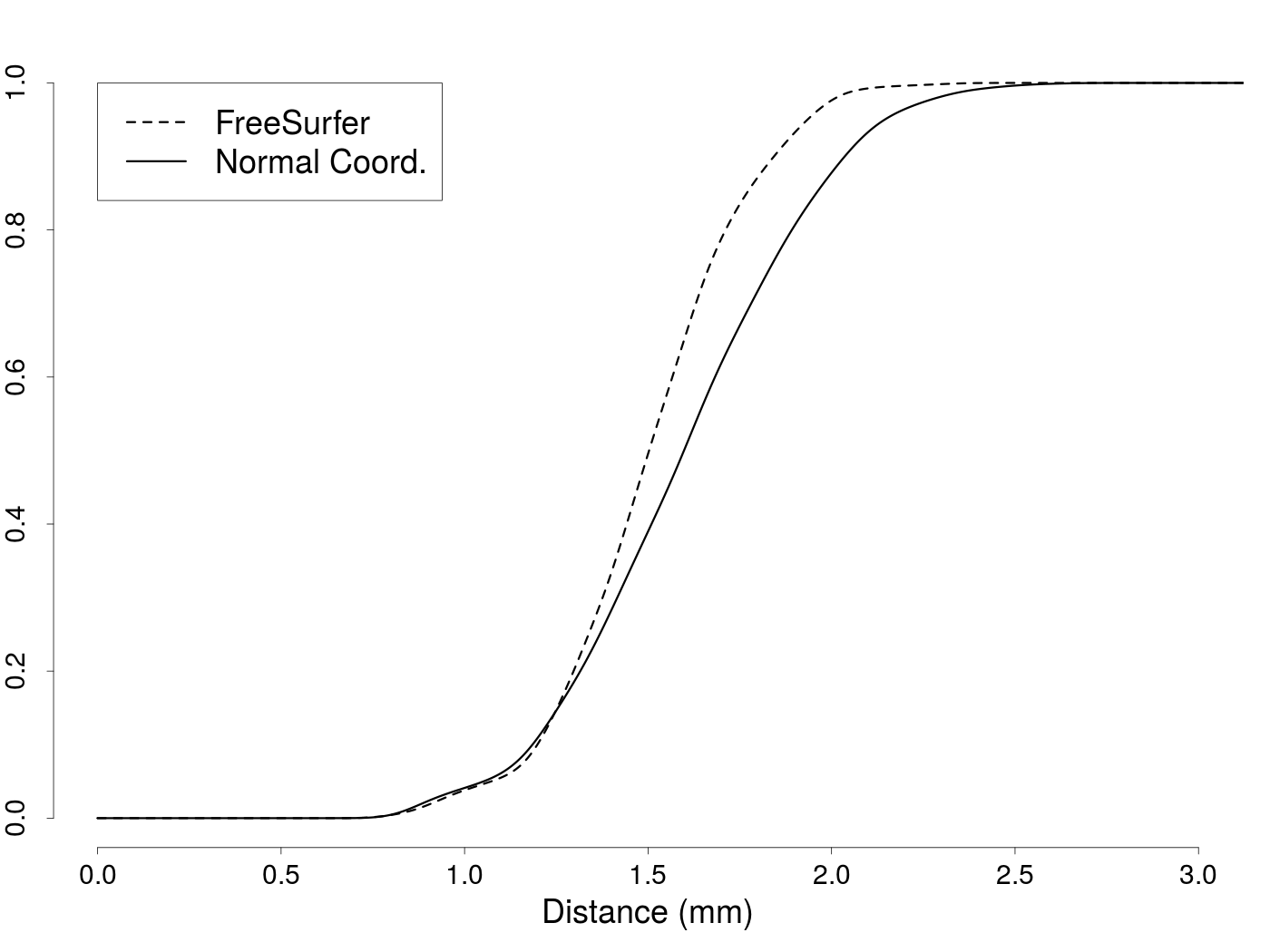} &
\includegraphics[width=0.45\textwidth]{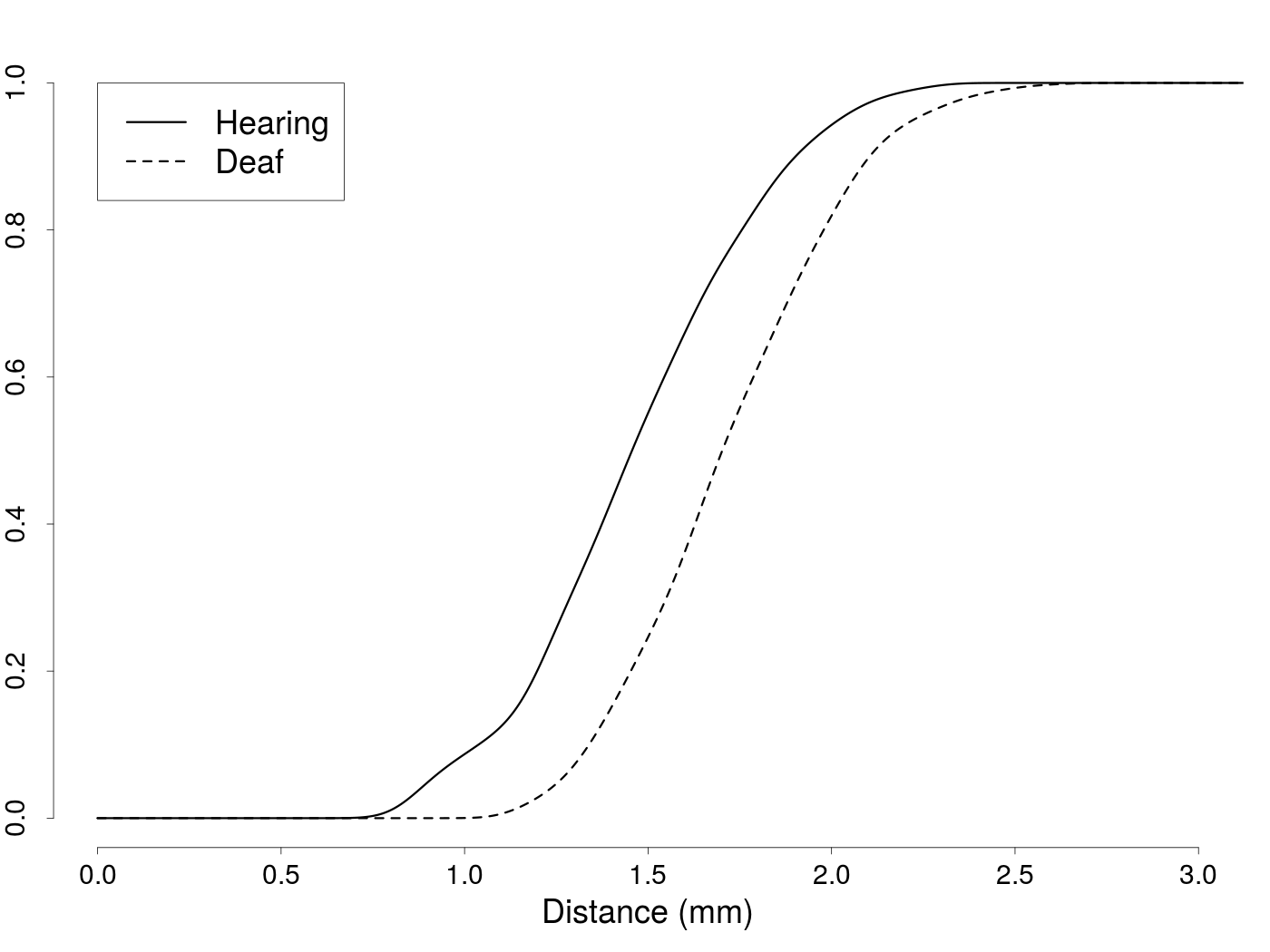}
\end{tabular}
\caption{Left: FreeSurfer distances underestimates those computed from the normal lines. Right: distances of normal lines in the congenitally deaf cat are bigger than those in the normal hearing cat.\label{Distances}}
\end{figure}

Since the lengths of the normal lines provide a notion of thickness, it is constructive to compare with those computed by widely-used brain mapping software, specifically FreeSurfer \citep{Fischl2012}. Given inner and outer surfaces $S^I$ and $S^O$, the distribution of distances with respect to all vertices (denoted by $r$ subscripted by the vertex index) of $S^I$ computed by FreeSurfer is
\begin{displaymath}
\left\{
\frac{\rho(r_i,f(r_i))+\rho(f(r_i),g(f(r_i)))}{2}
\forall r_i \in S^I
\right\}
\end{displaymath}
where $f(r_i) = \argmin\limits_{r_l \in S^O} | r_i - r_l |$,
$g(f(r_i)) = \argmin\limits_{r_l \in S^I} | f(r_i) - r_l |$
and $\rho(r_a,r_b) = ||r_a-r_b||$ is the Euclidean distance. Figure \ref{Distances} shows that FreeSurfer underestimates those computed by the present method and that the distribution of distances of the normal lines computed in Fig. \ref{Cats} for the deaf cat is different from that of the hearing cat.

\begin{figure}
\centering
\begin{tabular}{cc}
\includegraphics[width=0.47\textwidth]{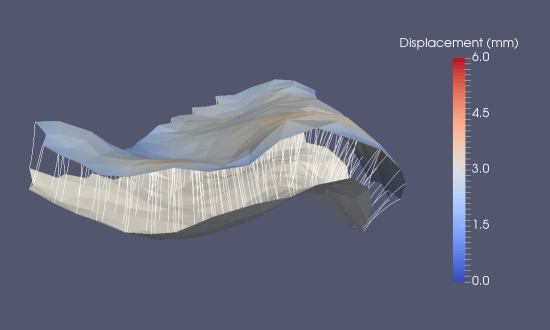} & \includegraphics[width=0.47\textwidth]{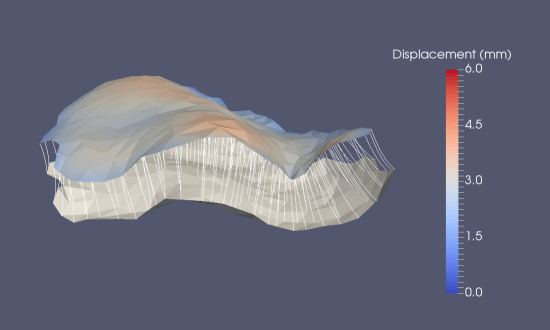} \\
\includegraphics[width=0.47\textwidth]{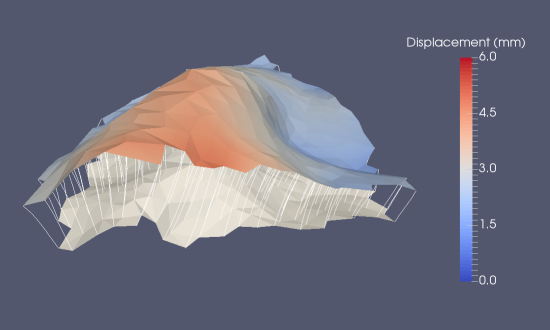} & \includegraphics[width=0.47\textwidth]{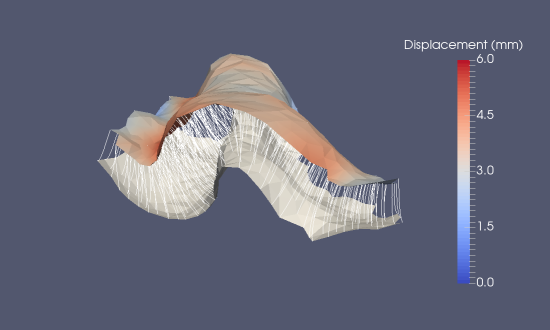}
\end{tabular}
\caption{Left and right planum temporale in a subject with normal hearing (top) and hearing loss (bottom).\label{HGPT}}
\end{figure}

\begin{figure}
\centering
\includegraphics[width=0.95\textwidth]{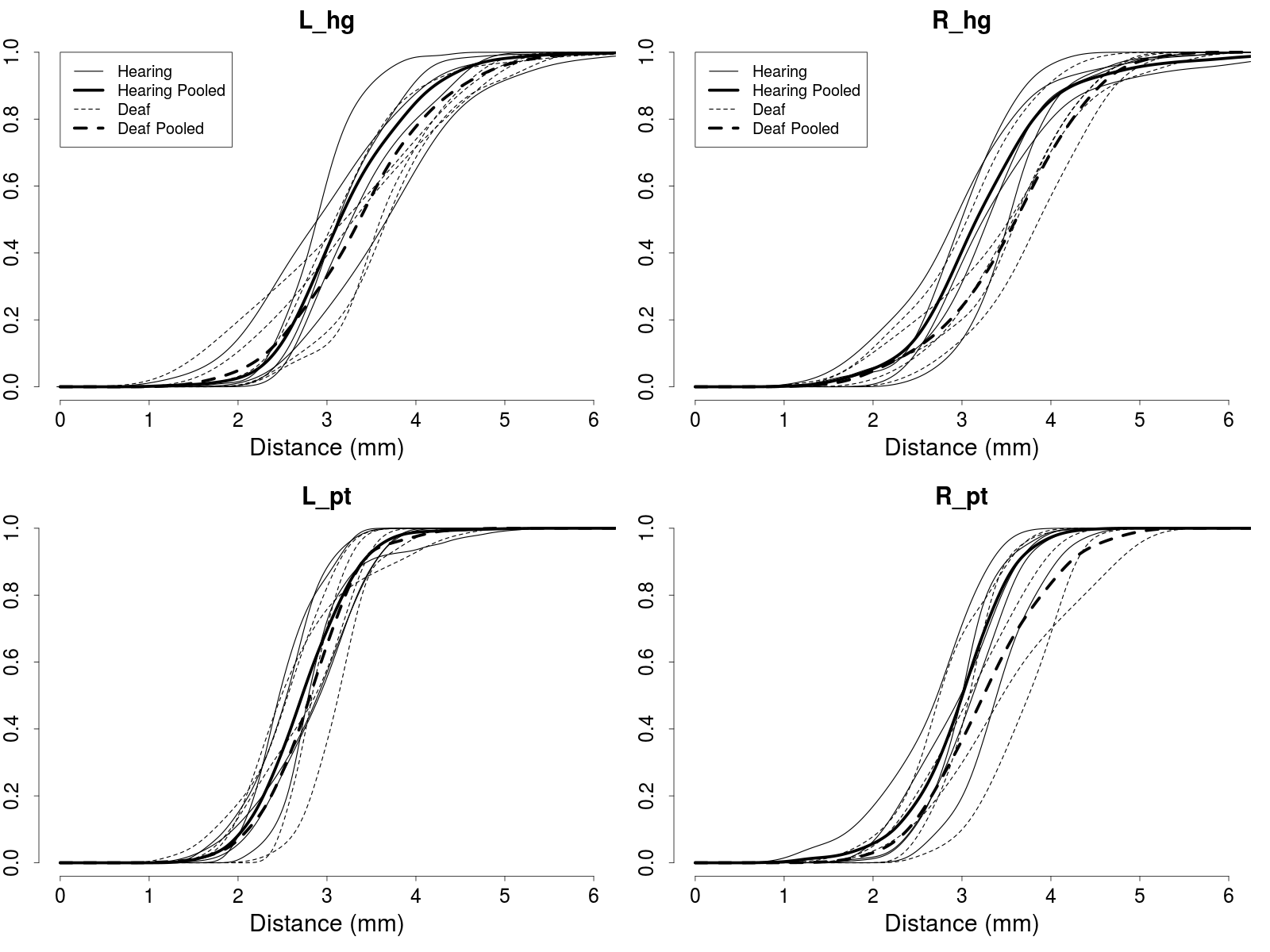}
\caption{Distances in the left and right Heschl's gyrus (top) and Planum temporale (bottom) in 10 adults (5 with profound hearing loss)\label{Adults}}
\end{figure}

Figure \ref{HGPT} shows normal lines for the left and right planum temporale in two adults - one with normal hearing and one with hearing loss. Figure \ref{Adults} shows the distribution of distances; also shown are pooled distributions. Table \ref{perf} shows the performance of typical computations for different structures.

\begin{table}
\centering
\tbl{\label{perf}Performance of typical computations}
{\begin{tabular}{lcccccc}
& \multicolumn{2}{c}{Inner} & \multicolumn{2}{c}{Outer} & & Runtime \\
\raisebox{1.5ex}[0pt]{Cortex}  &  Vertices & Faces & Vertices & Faces & \raisebox{1.5ex}[0pt]{Iterations}& (hours) \\
M1  & 239 & 417 & 375 & 670 & 2417 & 1.45 \\
Cat & 4178 & 8011 & 4401 & 8396 &  606 & 24.53 \\
PT  & 199 & 343 & 215 & 373 & 2343 & 1.17 \\
\end{tabular}}
\end{table}

\section{Discussion}
A columnar like coordinate system for cortical regions has been developed and applied in several cortices. The coordinates consists of the manifold evolving from the inner one to the outer one over $t\in[0,1]$. But it will be necessary to use histological data to ascertain if the evolving manifold (surface) follows the laminar properties of the cortical layers \citep{Waehnert2014,Leprince2015,Huntenburg2017,Bok1959}.

The approach adopted here exploits the geometric properties to generate the columnar like normal lines particularly in highly curved areas such as the gyral crowns and sulcal fundi which has been problematic in image-based approaches \citep{Leprince2015,Waehnert2014,Jones2000,Das2009} that also enforce orthogonality and equivolumetric laminar properties. In contrast, the triangulated inner surface forms a natural chart that is able to accommodate deeply buried sulci. For whole brain data, it may be easier to derive thickness from these image-based approaches but locally geometric based approaches such as the one described here may yield more reliable values. Even so, the method is computationally intensive in part due to using augmented Lagrangian to solve the optimal control problem \citep{arguillere2014shape}. It is not surprising that the run times are affected by number of faces and curvature. Speeding up computations could be achieved via a GPU or memory-efficient implementation of automatic differentiation libraries\cite{charlierkeops,tward2017unbiased,kuhnel2017computational}.  Thus this approach may be more suitable for cortical morphometry of regions of interest. So it would be interesting to evaluate the robustness of thickness computation with respect to different partitions of a larger area. Also an inexact matching method is used which means that the deformed inner surface does not exactly align with the target outer surface. There is therefore a small error in computing thickness due to the  misalignment which could be corrected by adjusting the normal lines at their crossing points with the outer surface. Finally, normal lines from open boundaries can be distorted due to the irregular boundaries caused by the triangulation. Strategies to address this include expanding outside cortical areas or imposing additional normal constraints.

It is clear that cortical areas have distributions of thickness due to curvature with sulci and gyri having smaller and larger distances respectively. Such distributions are amenable to statistical analysis not too dissimilar to previous work on labeled cortical depth maps \citep{Ceyhan2010}. It is not surprising given that distances generated by FreeSurfer underestimates those generated here.

With respect to the primary and secondary auditory areas, the deaf cat is thought to be thicker than the normal hearing cat \citep{Barone2013}. This is in contrast to the recent measurements of thickness based on histological data of gyral crowns \citep{Berger2017} which excludes the folds shown here. Also, differences in the myelination of axonal fibers in the neighborhood of the gray/white matter interface could influence results \citep{Emmorey2003}.

There are limited opportunities for studying adults with hearing loss who have been using listening and spoken language (LSL) via hearing aids since infancy \cite{Olulade2014}. Such people are difficult to recruit primarily because many now have cochlear implants. Yet they are likely to have normal (or near-normal) P1 and N1 latencies (from cortical auditory evoked potential data) which correlate with neural activity associated with primary and secondary auditory cortices i.e. Heschl's gyrus and Planum temporale \citep{Sharma2005,Kral2012,Sharma2015,Liem2012}. The pooled distributions suggest little differences in the left side which is associated with temporal processing \citep{Marie2016} and some differences on the right which is associated with spectral processing \citep{Marie2016}. The former may be attributed to auditory training used in LSL after early detection and intervention as infants while the latter may be attributed to high frequency hearing loss. By comparison, thicker visual cortical areas have been observed in people blinded since infancy \citep{Jiang2009}.


Future work will be focused initially on developing optimal sampling to build realistic 3D coordinates in cortical areas. In turn, these will be used for functional and stereological analysis particularly to assess the contributions from the laminar cortical layers to the overall properties of the cortical area.

\section*{Acknowledgements}
Funding from the National Institutes of Health (P41-EB015909, R01-DC016784), Federal Ministry of Education and Research (BMBF) of Germany (01GQ1703), National Organization for Hearing Research for the scans of adults with hearing loss, Institute of Mathematical Sciences at the National University of Singapore, the Isaac Newton Institute for Mathematical Sciences at the University of Cambridge (EP/K032208/1) during the Growth Form and Self-organisation programme are gratefully acknowledged. 

\bibliographystyle{./ws-rv-van}
\bibliography{cortex}

\end{document}